\documentclass[a4paper,11pt]{article}

\usepackage{fourier}
\usepackage{color}
\usepackage{graphicx}
\usepackage{url}
\usepackage[affil-it]{authblk}
\usepackage{amsmath}
\usepackage{wrapfig}
\usepackage{siunitx}
\usepackage[hidelinks]{hyperref}

\usepackage{booktabs,dcolumn,caption}

\newcolumntype{d}[1]{D{.}{.}{#1}} 

\usepackage{amssymb}
\usepackage{pifont}
\newcommand{\cmark}{\ding{51}}%
\newcommand{\xmark}{\ding{55}}%

\usepackage[T1]{fontenc}
\usepackage{times}

\usepackage[x11names]{xcolor}

\usepackage[flushleft]{threeparttable}
\usepackage{float}

\begin{document}

\title{Velocity Driven Vision: Asynchronous Sensor Fusion Birds Eye View Models for Autonomous Vehicles}
\author{Seamie Hayes, Sushil Sharma, Ciarán Eising}
\affil{University of Limerick, Ireland}

\affil{\bf This paper is a preprint of a paper submitted to the 26th Irish Machine Vision and Image Processing
Conference (IMVIP 2024). If accepted, the copy of record will be available at IET Digital Library.}
\date{}
\maketitle
\thispagestyle{empty}

\ifx00 
\newcommand{\SKS}[1]{\textcolor{red}{[Sushil: #1]}}
\else 
\newcommand{\SKS}[1]{\textcolor{red}{}}
\fi

\begin{abstract}
Fusing different sensor modalities can be a difficult task, particularly if they are asynchronous. Asynchronisation may arise due to long processing times or improper synchronisation during calibration, and there must exist a way to still utilise this previous information for the purpose of safe driving, and object detection in ego vehicle/ multi-agent trajectory prediction. Difficulties arise in the fact that the sensor modalities have captured information at different times and also at different positions in space. Therefore, they are not spatially nor temporally aligned. This paper will investigate the challenge of radar and LiDAR sensors being asynchronous relative to the camera sensors, for various time latencies. The spatial alignment will be resolved before lifting into BEV space via the transformation of the radar/LiDAR point clouds into the new ego frame coordinate system. Only after this can we concatenate the radar/LiDAR point cloud and lifted camera features. Temporal alignment will be remedied for radar data only, we will implement a novel method of inferring the future radar point positions using the velocity information. Our approach to resolving the issue of sensor asynchrony yields promising results. We demonstrate velocity information can drastically improve IoU for asynchronous datasets, as for a time latency of 360 milliseconds (ms), IoU improves from 49.54 to 53.63. Additionally, for a time latency of 550ms, the camera+radar (C+R) model outperforms the camera+LiDAR (C+L) model by 0.18 IoU. This is an advancement in utilising the often-neglected radar sensor modality, which is less favoured than LiDAR for autonomous driving purposes.
\end{abstract}

\section{Introduction}
The area of 3D perception is a growing area of research and is imperative in the applications of autonomous driving and computer vision. The dangers of driving are not something that can be disregarded, and there has been huge work devoted to making roads safer, such as the drink-driving ban in Ireland in 1969, along with the wearing of seatbelts becoming mandatory in 1979. With the recent digital revolution of the past decades, particularly the huge breakthroughs in machine learning, extensive work has been put into applying these to areas of driving to aid the driver in difficult scenarios or even fully autonomous driving where there is no driver present. One of the major hurdles of autonomous driving is the idea of 3D perception. How is it possible to provide the vehicle information so that it can efficiently process it and make quicker decisions than a human?

\vspace{0.25em}
The mainstream method to acquire data to feed to a network for 3D perception is via three sensors: camera, radar, and LiDAR. These sensors can be used alone or together, a method known as multi-modal fusion. Camera sensors provide a flat 2D representation of the surroundings but have the advantage of supplying us with colour information, which allows us to discern road markings and stop signs. Radar and LiDAR provide us with a 3D representation of the world in a point cloud where these sensors emit signals that will hit objects and reflect back to the sensor, indicating the presence of an object. One major advantage LiDAR provides is the density of its point cloud \cite{radsparse}. In the widely known nuScenes dataset, LiDAR data is approximately 40 times denser than radar data \cite{nuScenes}. This density lends itself to greater model performance, which is a huge reason why there has been much more research in the area of C+L fusion. 

\vspace{0.25em}
However, radar has a distinct capability that we will utilise in this research, that being the velocity information. The nuScenes vehicle, known as the ego vehicle, is equipped with five Continental ARS 408-21 radar sensors, which are capable of capturing velocity information in one measurement cycle. However, the LiDAR sensor used, Velodyne HDL 32 E, is incapable of such. This velocity information can be utilised in a novel way to predict the future position of these radar points, which as we will see in the results section, gives a notable boost in performance in asynchronous C+R models. This boost in the performance of asynchronous C+R models places them ever closer to the performance of asynchronous C+L models. Previous research has struggled to find use cases where C+R models can outperform C+L models.

\begin{wrapfigure}{r}{0.45\textwidth} 
  \vspace{-20pt} 
  \begin{center}
    \includegraphics[width=0.5\textwidth]{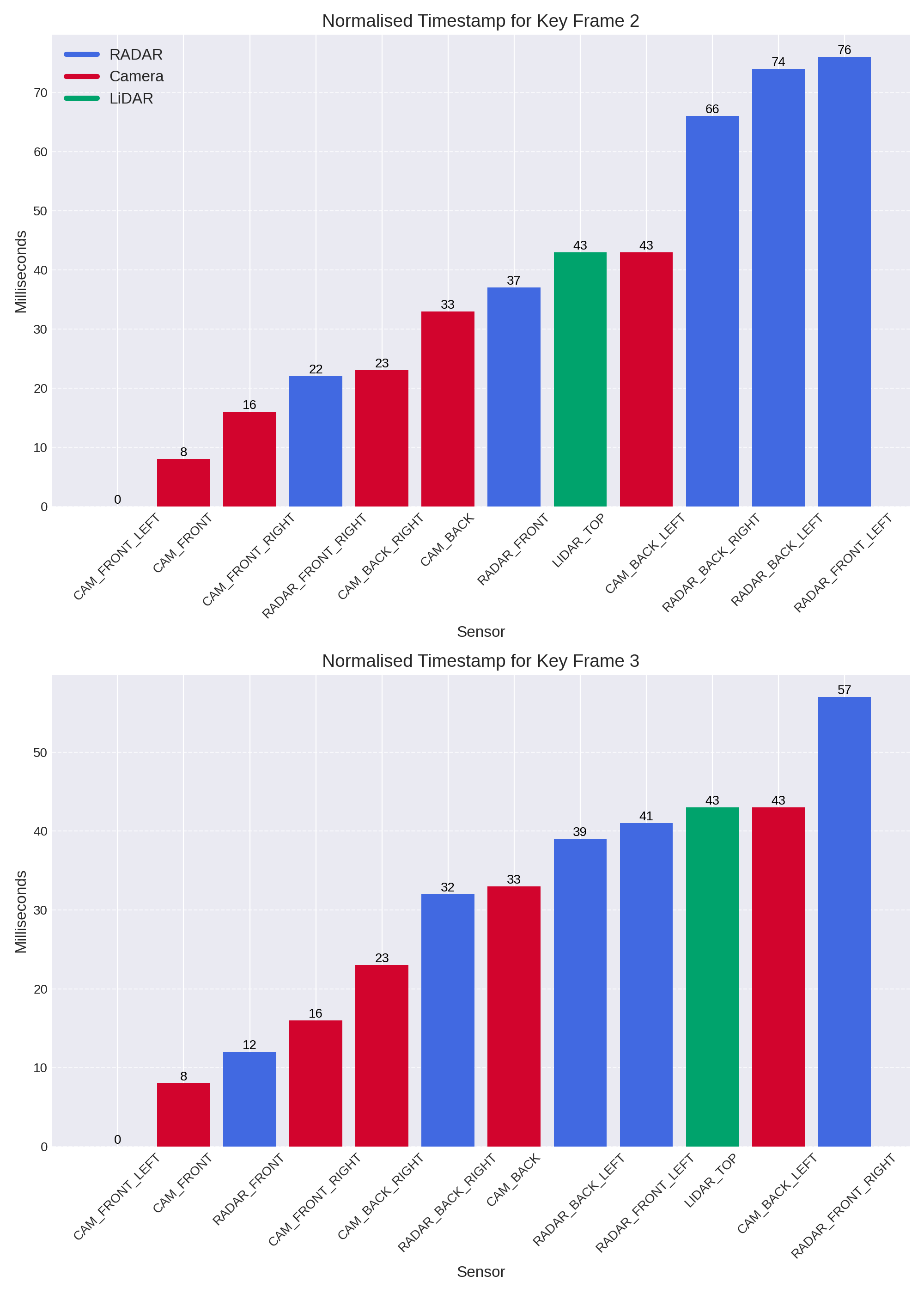}
  \end{center}
  \caption{Millisecond timestamp capture of all sensors for two consecutive keyframes. Time is standardised such that \texttt{CAM\_FRONT\_LEFT's }timestamp is zero}
  \label{timestamp}
\end{wrapfigure}

\vspace{0.25em}
One aspect of the nuScenes dataset that must be noted is the radar sensors are not synchronised with the cameras, whereas the LiDAR sensors are synchronised with the cameras. This is evident in Figure \ref{timestamp} where we can see that for two consecutive frames, LiDAR and camera timestamps are consistent, whereas radar is sporadic. This is a limitation of radar sensors in the nuScenes dataset and will introduce problems with spatial and temporal alignment in perception algorithms. This paper aims to examine the limit of asynchrony via manual manipulation of the nuScenes dataset, which may occur due to incorrect calibration of sensors. We will analyse this problem in the context of Birds Eye View (BEV) Transformers \cite{simplebev}. The BEV space, put very simply, is the 3D space with the $z-axis$ flattened to give a 2D plane, which is a more simple representation of the world. Due to this, it has been widely used, and well researched \cite{shi2023gridcentric}.

\vspace{0.25em}
In short, this paper aims to improve the safety of autonomous vehicles via a comprehensive exploration of asynchronous radar and LiDAR sensors. 
The main contributions are as follows:

\begin{enumerate}
  \item An investigation into the effect asynchrony has on radar and LiDAR sensors via the construction of a novel asynchronous and corresponding synchronous dataset made from the nuScenes dataset.
  \item A novel implementation of radar velocity information to infer future radar points for asynchronous radar sensors. For a time latency of 360ms, we examine IoU improvement from 49.54 to 53.63
\end{enumerate}

\section{Background}
\subsection{State of the Art}
Following vehicle segmentation, trajectory prediction of both our ego vehicle and other vehicles is a crucial and extensively studied task \cite{hu2021fieryfutureinstanceprediction, Sharma2024BEVSeg2TP}. Accurately detecting vehicle position and orientation is crucial for precise movement prediction. Our approach to resolving asynchrony demonstrates promising results in addressing this challenge.

\vspace{0.25em}
The task of lifting camera images into 3D or BEV space can be achieved in numerous ways. 2D features can be `pushed' from the image into 3D \cite{liftsplat, bevcar}, whereas some methods opt to pull features from the image to each 3D voxel \cite{simplebev} which is the method we will utilise for our experiments. 

\begin{wraptable}{r}{0.25\textwidth}  
\begin{tabular}{ *{2}{c} } 
    \toprule
    \multicolumn{1}{c}{\textbf{Modality}} &\multicolumn{1}{c}{\textbf{IoU}} \\
        \midrule
    Camera & 47.4 \\
    Camera + radar & 55.7 \\
    Camera + LiDAR & 60.8 \\
    \bottomrule
  \end{tabular}
    \caption{Multi-Modal Fusion Analysis \cite{simplebev}}
    \label{simplebev_results}
\end{wraptable}

\vspace{0.25em}
Multi-modal fusion is an area of research with a large emphasis placed on it, namely due to the complementary nature both radar, and LiDAR have on cameras due to depth perception. C+L fusion provides huge improvements over camera-only \cite{bevformer} and LiDAR-only models \cite{bevfusion,deepfusion}. However, LiDAR sensors can be unattractive as they are quite costly. C+R fusion has been seen to give great improvement over single modality models using cameras only \cite{bevcar, simplebev}. Table \ref{simplebev_results} displays the benefits of multi-modal fusion. In the field of multi-modal fusion, the issue of asynchrony has been largely neglected. Particularly in the context of C+R fusion, one would expect this to be a focal point of discussion, especially given the inherent asynchrony present in the nuScenes dataset. Here, we aim to address and resolve the challenge of asynchronisation.

\subsection{nuScenes Dataset}
\begin{wrapfigure}[14]{r}{0.4\textwidth} 
  \vspace{-30pt} 
  \begin{center}
    \includegraphics[width=0.4\textwidth]{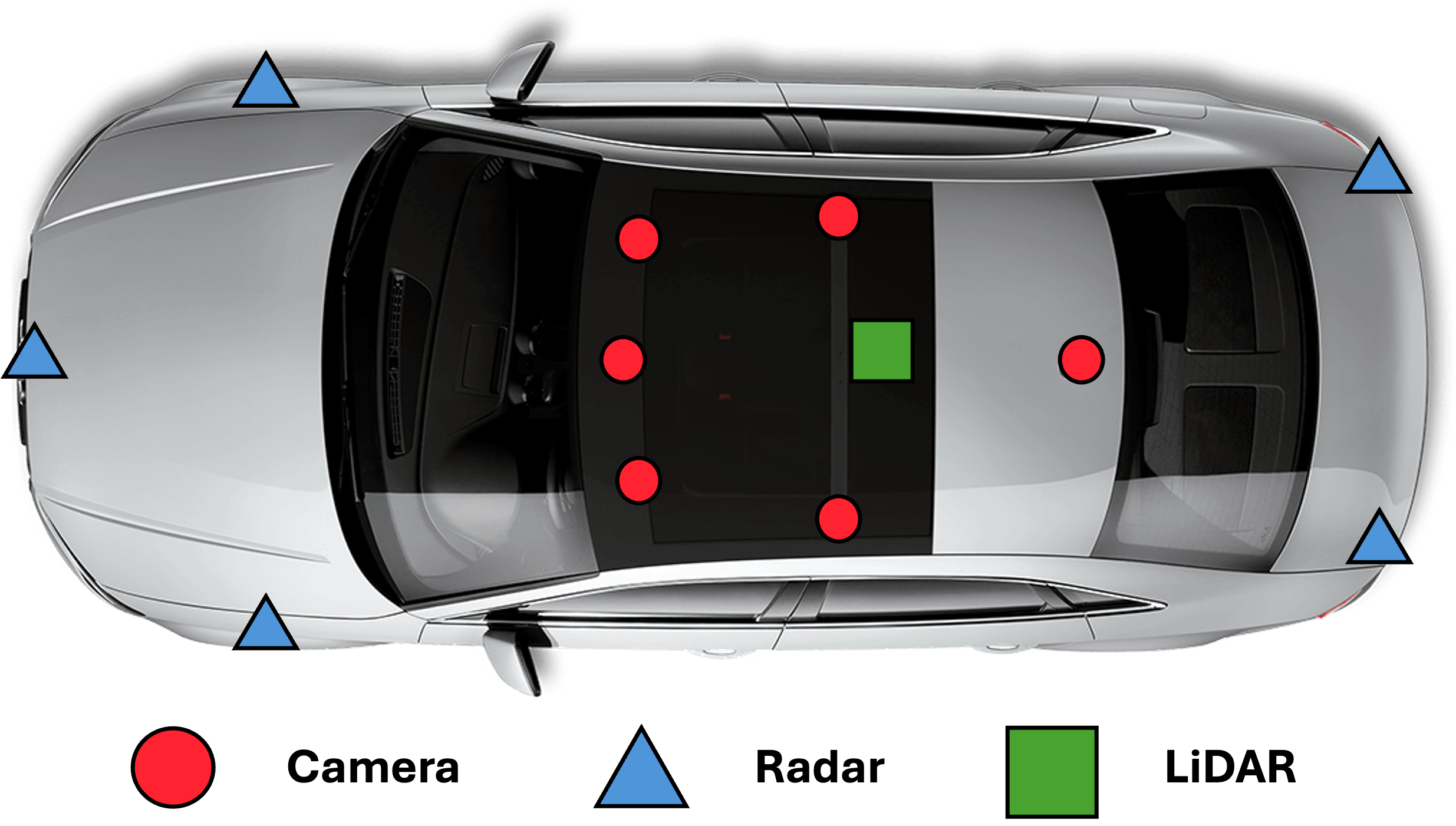}
  \end{center}
  \caption{Top-down view of a rendition of the nuScenes ego vehicle with its sensor modalities annotated}
  \label{nusc_car}
\end{wrapfigure}

For the purpose of these experiments, we utilise the widely known nuScenes dataset \cite{nuScenes}, namely due to its numerous modality sensors and extensive amount of data available for training.

\vspace{0.25em}
The vehicle used to capture the nuScenes dataset is equipped with six cameras, five radar sensors, and one LiDAR sensor as seen in Figure \ref{nusc_car}. LiDAR operates at 20Hz, performing a 360-degree sweep from the front left camera to the back left camera. Cameras are synchronized to LiDAR, as each camera is triggered once the LiDAR sweep covers its field of view. Consequently, the front left camera is triggered first, and the timing and order of captures remain constant for all keyframes (Figure \ref{timestamp}). In contrast, radar is captured at a rate of 13Hz \cite{nuScenes}, with a non-constant ordering of capture across keyframes making it inherently asynchronous, although only slightly. To further explore the boundaries of asynchrony, we will examine greater time latencies, specifically considering latencies exceeding 500 milliseconds.

\subsection{Simple-BEV}

\begin{figure*}[h!]
  \begin{center}
    \includegraphics[width=1\textwidth]{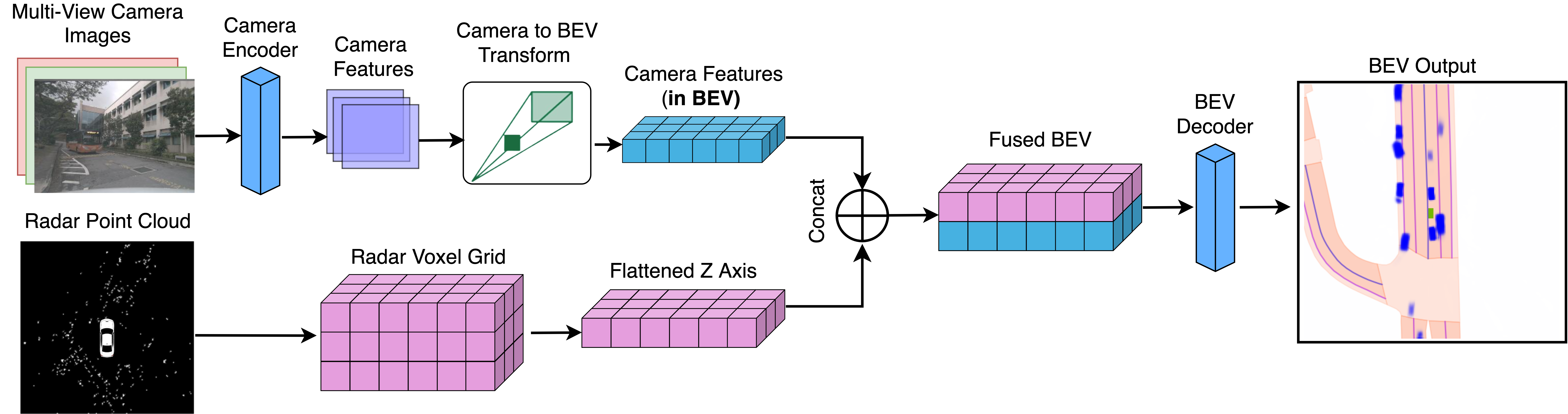}
  \end{center}
  \captionsetup{justification=centering}
  \caption{Architecture flow diagram for Simple-BEV}
  \label{arch}
\end{figure*}

The machine learning model used for these experiments is based on Simple-BEV \cite{simplebev}. Firstly, two 3D voxel grids centred on the chosen reference camera of size $200\times200\times8$ are initialized for both camera and radar \cite{simplebev}. A 2D ResNet-101 backbone is applied to each camera image to extract features such as pedestrians, vehicles, etc. Following this, bilinear sampling is performed to lift these features into the 3D voxel which was initialized previously, and this is then flattened to yield camera BEV features as illustrated in Figure \ref{arch}. The radar point cloud occupies the remaining voxel grid. The voxel grids are concatenated to achieve a unified voxel grid, which is then flattened on the $z$ axis to achieve a BEV grid. The unified voxel grid is input to the BEV decoder, producing our BEV vehicle segmentation output.

\vspace{0.25em}
We evaluate our C+R datasets using the pre-trained C+R model provided. We must train our C+L model independently as it is not provided, and we employ the same hyperparameters as those used for the C+R model.

\section{Methodology}
\subsection{Construction of the Asynchronous and Synchronous Datasets}

\begin{wrapfigure}{r}{0.6\textwidth} 
  \begin{center}
    \includegraphics[width=0.6\textwidth]{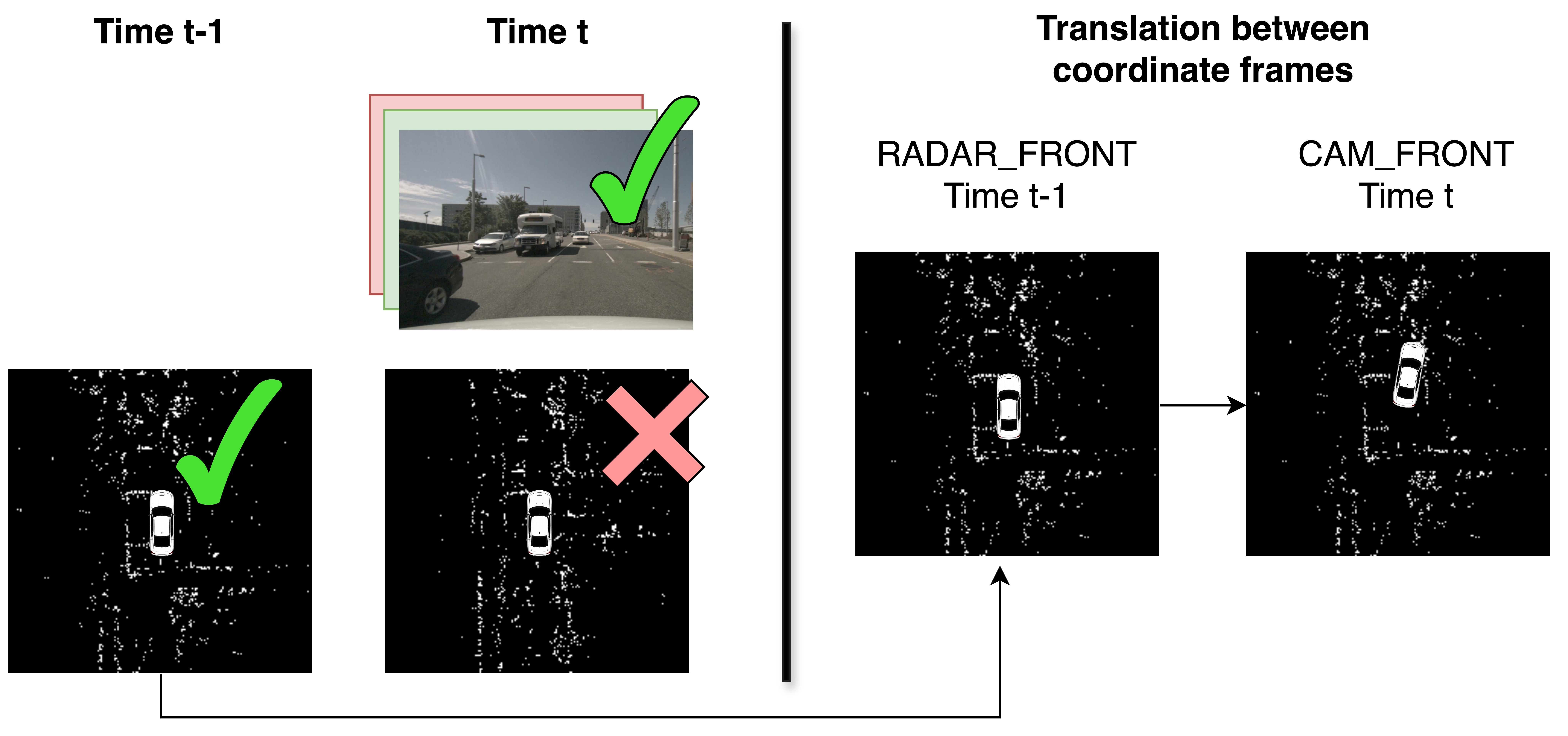}
  \end{center}
  \caption{\textbf{Asynchronous Dataset:} Construction and translation between coordinate frames}
  \label{async_dataset}
\end{wrapfigure}

Each driving scene in the nuScenes dataset is comprised of keyframes which are annotated with ground truth data for the purpose of supervised training. In between these keyframes are additional data captures that are not annotated. We will utilize this additional data to make our dataset asynchronous. Manipulation of the data will yield two new datasets which we will refer to as the \textit{synchronous}, and \textit{asynchronous} datasets. The synchronous dataset will serve as our baseline for comparing the asynchronous dataset to analyze the degradation. For the synchronous dataset, we remove the first two keyframes of data to ensure the consistency of previous radar/LiDAR sweeps between both datasets.

\vspace{0.25em}
For the asynchronous dataset, we again remove the first two keyframes. We then take the radar/LiDAR data from time $t-1$ as seen in Figure \ref{async_dataset}. Given a time latency of approximately 140 ms for radar, this involves using the radar data two captures prior. One challenge in this process is that the radar/LiDAR data is still stored in reference to its old ego-vehicle frame. We must translate it into the current ego frame, and this is done via a straightforward transformation between coordinate systems:
\begin{align*}
 \mathsf{EgoFrame}_{\mathsf{ \ t-1}} \rightarrow \mathsf{Global} \rightarrow \mathsf{EgoFrame}_{\mathsf{ \ t}}  
\end{align*}

One important note is that we assume we do not have the ego frame for radar for current time $t$. Instead, we will translate the radar/LiDAR data to the ego frame of \texttt{CAM\_FRONT} as illustrated in Figure \ref{async_dataset}. The reasoning for choosing \texttt{CAM\_FRONT} is this is the reference camera during validation.

\vspace{0.25em}
Thus two datasets are constructed. As mentioned previously, radar and camera are not synchronous, the term synchronous dataset is not to be taken literally and is used for distinction purposes. We will train the model displayed in Figure \ref{arch} using the above-constructed datasets. Time latencies greater than 360ms yield similar results and are therefore excluded for the sake of brevity. C+L datasets are constructed in a manner that approximates the same time latency as an equivalent C+R closely as possible for fair comparison

\begin{wrapfigure}[16]{r}{0.5\textwidth}  
  \vspace{-24pt} 
  \begin{center}
    \includegraphics[width=0.5\textwidth]{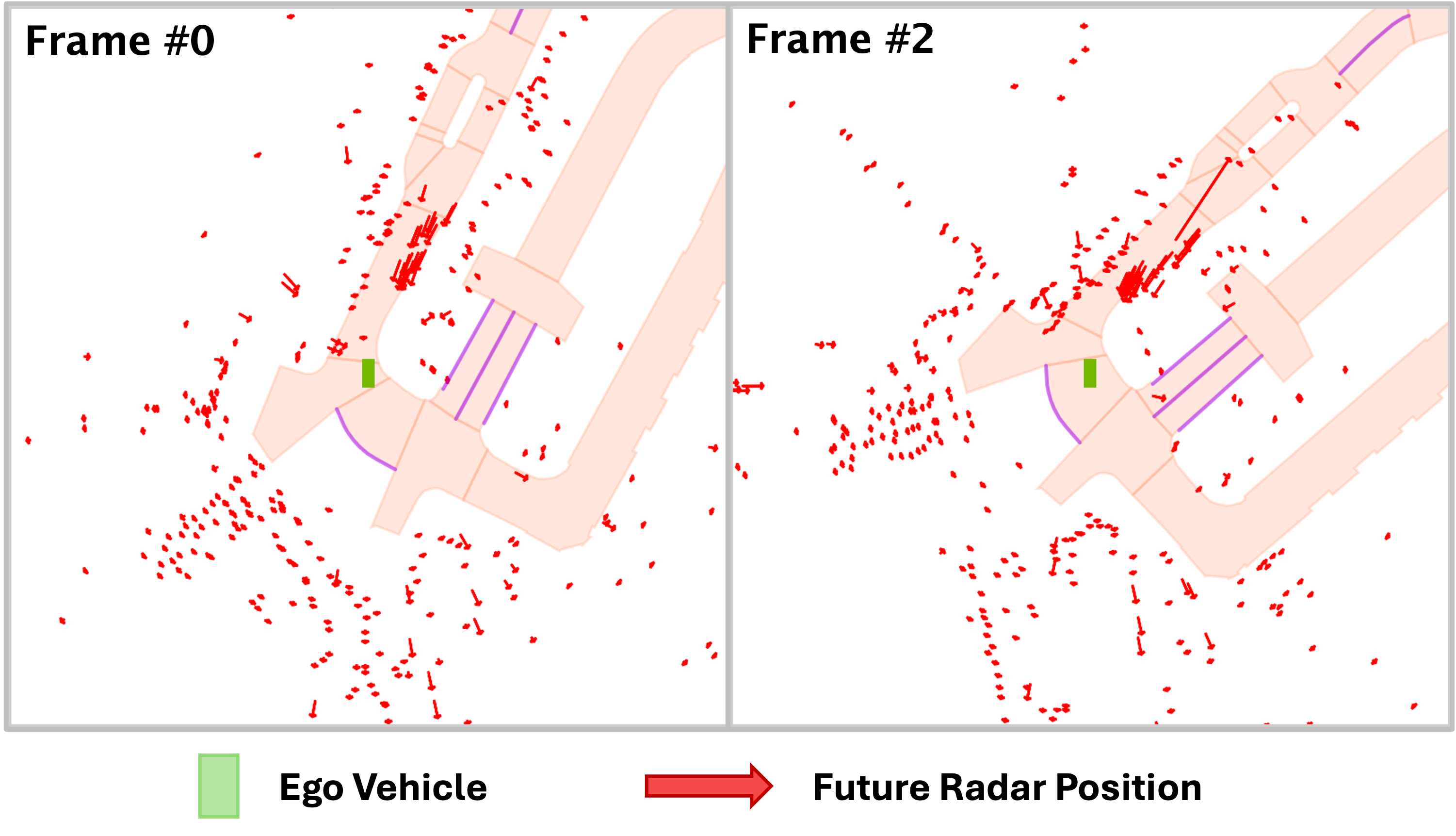}
  \end{center}
  \caption{\textbf{Visualization of inferring future radar points}: The red arrow indicates the movement of the original radar point to its inferred future position}
  \label{radar_visual}
\end{wrapfigure}

\subsection{Enhancing Radar Data with Velocity Information}
As mentioned previously, the radar sensors equipped on the ego vehicle are capable of capturing velocity information, with an accuracy of ±0.1 km/h \cite{nuScenes}. Therefore, for our outdated radar data at time $t-1$, we can update the point positions, $P_{t-1}$, by inferring what their current position should be, $P_{t}$, with the $xy$ velocity information, $P_{vel}$. Note, that there is no vertical ($z$) velocity. As noted previously, we represent our data relative to the ego frame of \texttt{CAM\_FRONT}, and so the duration of time which has passed is the timestamp of \texttt{CAM\_FRONT} $t_{cam}$, minus the timestamp our of radar sensor, $t_{radar}$. We use the following formula to predict future position:
\begin{align*} 
P_{t} \ = \ P_{t-1} \ + \ P_{vel} \times [t_{cam} - t_{radar}]
\end{align*}

\section{Results}
\subsection{Quantitative Analysis: Intersection over Union}
The first step of evaluating the performance of our models is qualitative analysis which we will do using the IoU metric. IoU refers to Intersection over Union and is a widely used metric for evaluating object detection models. It is computed by taking the predicted bounding boxes $\text{B}_\text{pred}$ from the BEV decoder seen in Figure \ref{arch} and comparing them to the ground truth bounding boxes $\text{B}_\text{gt}$. This is calculated using the well-known IoU formula:
\begin{gather*}
\text{IoU} = \frac{|\text{B}_{\text{gt}} \cap \text{B}_{\text{pred}}|}{|\text{B}_{\text{gt}} \cup \text{B}_{\text{pred}}|}
\end{gather*}

One important note is that IoU penalizes the model in the case of predicting a vehicle in a position where there is none, which is crucial for ensuring safety in autonomous driving systems.

\begin{table}[!ht]
\centering
\renewcommand{\arraystretch}{1} 
\begin{tabular}{*{9}{c}}
    \toprule
    \multicolumn{4}{c}{\textbf{Configuration}} &\multicolumn{2}{c}{\textbf{Metrics}} \\
    \cmidrule(lr){1-4}\cmidrule(lr){5-6}
\textbf{\textit{Sync/Async}} & \textbf{\textit{Modality}} & \textbf{\textit{Latency}}  & \textbf{\textit{Infer Position}} & \textbf{\textit{IOU} $\uparrow$} & \textbf{\textit{Velocity Improvement}}  \\
\hline
\hline
Sync       & C+R      & 0ms  & \cmark   & \textbf{55.60}       & \textbf{\textcolor{RoyalBlue3}{+0.03}}                       \\ 
Sync       & C+R      & 0ms  & \xmark    & \textbf{55.57}       &                                          \\ \hline
Async      & C+R      & 70ms & \cmark   & \textbf{55.33 \textcolor{Red2}{(-0.27)}}        & \textbf{\textcolor{RoyalBlue3}{+0.69 }}                      \\ 
Async      & C+R      & 70ms & \xmark    & \textbf{54.64 \textcolor{Red2}{(-0.93)}}            &                                    \\ \hline
Async      & C+R      & 140ms & \cmark  & \textbf{54.91  \textcolor{Red2}{(-0.69)}}              &\textbf{\textcolor{RoyalBlue3}{+1.59 } }                            \\ 
Async      & C+R      & 140ms & \xmark   & \textbf{53.32  \textcolor{Red2}{(-2.25)}}              &                                     \\ \hline
Async      & C+R      & 220ms & \cmark  & \textbf{54.49  \textcolor{Red2}{(-1.11)}}              &\textbf{\textcolor{RoyalBlue3}{+2.55 } }                               \\ 
Async      & C+R      & 220ms & \xmark   & \textbf{51.94   \textcolor{Red2}{(-3.63)} }            &                                    \\ \hline
Async      & C+R      & 290ms & \cmark  & \textbf{54.01  \textcolor{Red2}{(-1.59}}              & \textbf{\textcolor{RoyalBlue3}{+3.32 }}                              \\ 
Async      & C+R      & 290ms & \xmark   & \textbf{50.69   \textcolor{Red2}{(-4.88)}}             &                                    \\ \hline
Async      & C+R      & 360ms & \cmark  & \textbf{53.63   \textcolor{Red2}{(-1.97)}}             &\textbf{ \textcolor{RoyalBlue3}{+4.09 } }                             \\ 
Async      & C+R      & 360ms & \xmark   & \textbf{49.54 \textcolor{Red2}{(-6.03)}}               &                                 \\ \hline \hline
Async      & C+R      & 570ms & \cmark  & \textbf{52.30   \textcolor{Red2}{(-3.30)}}             &\textbf{ \textcolor{RoyalBlue3}{+5.12 } }                             \\ 
Async      & C+R      & 570ms & \xmark   & \textbf{47.18 \textcolor{Red2}{(-8.39)}}               &                                 \\ \hline
\end{tabular}
\caption{\textbf{IoU Comparirson for various asynchronous C+R models}: Note the degradation asynchrony introduces (seen in \textcolor{Red2}{red}) and the improvement inferring position using velocity has on performance (seen in \textcolor{RoyalBlue3}{blue}). Asynchrony degradation is relative to its synchronous equivalent (inferring position).}
\label{tab:iou_cr}
\end{table}

\begin{table}[!ht]
\centering
\begin{tabular}{*{5}{c}}
    \toprule
    \multicolumn{3}{c}{\textbf{Configuration}} &\multicolumn{2}{c}{\textbf{Metrics}} \\
    \cmidrule(lr){1-3}\cmidrule(lr){4-5}
\textbf{\textit{Sync/Async}} & \textbf{\textit{Modality}} & \textbf{\textit{Latency}} & \textbf{\textit{IOU} $\uparrow$} & \textbf{\textit{Asychrony Degradation}} \\
\hline

Sync       & C+L      & 0ms    & \textbf{62.66}       &                     \\ \hline
Async      & C+L      & 50ms   & \textbf{62.07}       & \textbf{\textcolor{Red2}{-0.57} }              \\ 
Async      & C+L      & 150ms  & \textbf{60.01}       & \textbf{\textcolor{Red2}{-2.65}}              \\ 
Async      & C+L      & 200ms  & \textbf{58.86}       & \textbf{\textcolor{Red2}{-3.80}}              \\ 
Async      & C+L      & 300ms  & \textbf{56.57}       & \textbf{\textcolor{Red2}{-6.09} }              \\ 
Async      & C+L      & 350ms  & \textbf{55.52}       & \textbf{\textcolor{Red2}{-7.14}}              \\ \hline \hline
Async      & C+L      & 550ms  & \textbf{52.12}       & \textbf{\textcolor{Red2}{-10.54}}  \\ \hline
\end{tabular}
\caption{\textbf{IoU Comparirson for various asynchronous C+L models}: Asynchrony introduces a great amount of degradation (seen in \textcolor{Red2}{red}).}
\label{tab:iou_cl}
\end{table}

\vspace{0.25em}
First, we will analyze the impact of asynchrony on the performance of both C+R and C+L models' performance as shown in Table \ref{tab:iou_cr} and \ref{tab:iou_cl}. Analysis of the C+R asynchronous models reveals a minor performance decline when the radar data latency is 70ms, with an IoU drop of 0.93. This trend continues, where for greater time latencies we note a greater reduction in performance. For radar data latency 360ms we examine a massive 6.03 drop in IoU. Similar results can be observed for the C+L models where for a near equivalent 350ms latency, we see a 7.14 drop in IoU. It appears that as the data is becoming progressively outdated, the performance is tending towards a camera-only model. The more pronounced drop in IoU for C+L compared to C+R can be attributed to the significant performance boost LiDAR data provides initially.

\vspace{0.25em}
The impact of inferring future point positions using velocity information in C+R models cannot be understated. For the 70ms model, we observe a 0.69 increase in IoU, placing it only 0.27 IoU points below the equivalent synchronous model. Moreover, the gains in IoU for the inferring points model continue to increase as the radar data latency grows. At a latency of 360 ms, the difference between the asynchronous inferring model and the corresponding synchronous model's IoU is 1.97 points. However, as expected, the performance gap between the asynchronous and synchronous models widens with increasing time latency.

\vspace{0.25em}
Comparing the C+R models to the C+L models, it is evident the C+L model outperforms C+R by a significant margin. For the base synchronous model, the difference is 7.09 IoU. However, this gap narrows for the asynchronous models as radar velocity information is utilized to appreciable effect at greater time latencies. At a latency of 560ms, the C+R model surpasses the C+L model. Although this occurs at quite a large latency, it is a rare instance where a C+R model can outperform a C+L model.

\subsection{Qualitative Analysis: BEV Vehicle Segmentation Comparison}
For the qualitative analysis, we will visually inspect the output of the BEV encoder from select models against the ground truth for a selected key frame of a scene. The specific frame is chosen due to the abundance of moving vehicles in traffic.

\vspace{0.25em}
The impact of asynchrony is particularly evident in both the C+R and C+L models, especially with time latencies exceeding 500 milliseconds. In the 70ms latency dataset, we examine minor misalignments of vehicles, namely for the vehicles highlighted in red and pink. Notably, the vehicle highlighted in orange is not segmented in any of the ~500ms models.

\vspace{0.25em}
Inferring future radar point cloud positions significantly improves BEV vehicle segmentation, especially for the great latency of 560ms. Inferring future data does not necessarily allow the model to detect more vehicles, however, it positions them more accurately due to the correct alignment of radar points. This is most prominent for the vehicles highlighted by the red and pink boxes. This change is not prominent for the synchronous models, as reflected by a minimal IoU difference of only 0.03 seen in the previous analysis.

\vspace{0.25em}
Comparing the C+R and C+L models, it is evident that C+L excels in vehicle segmentation. Note that the vehicle highlighted in cyan is not segmented in any C+R model. In the previous analysis, we found that C+R with inferred positions outperforms the C+L model at a time latency of approximately 500ms. Although C+L can detect more vehicles in this case, such as the cyan vehicle, C+R provides better-aligned vehicle segmentation when inferring future data. This alignment is crucial for ego vehicle and multi-agent trajectory prediction models.

\begin{figure}[h!]
  \centering
  \makebox[\textwidth][c]{%
    \resizebox{1\textwidth}{!}{%
      \includegraphics{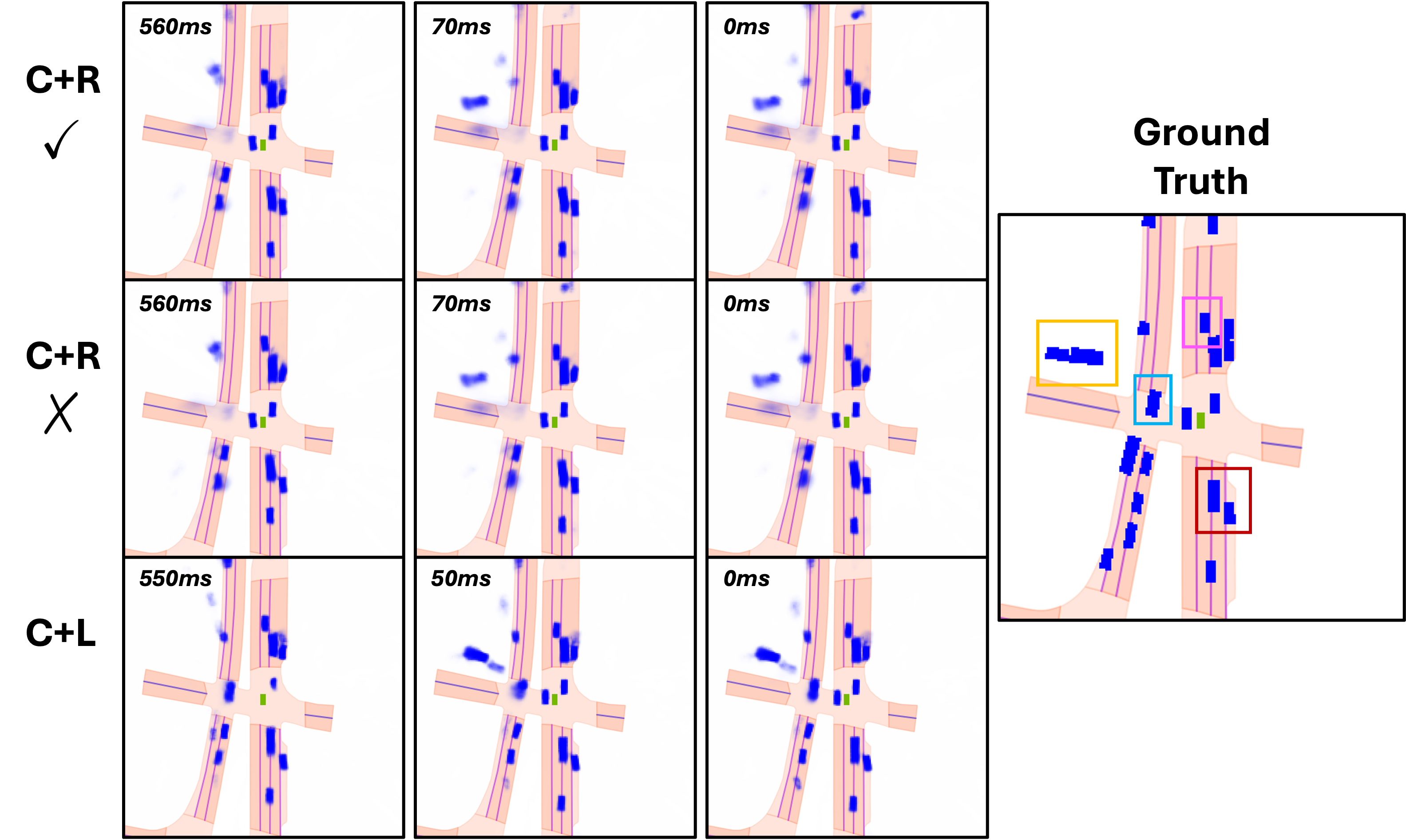}
    }%
  }
  \caption{\textbf{BEV image comparison for select experiments}: C+R tick involves inferring future radar point cloud positions.C+R cross indicates not inferring future positions. Time latency is labelled on each image. Some vehicles are surrounded by coloured boxes for analysis.}
  \label{bev_grid}
\end{figure}

\section{Conclusion}
To conclude, it is apparent that the asynchrony of radar and LiDAR severely degrades the performance of multi-modal models. For radar data, inferring future positions using velocity information provides a major boost in performance and approaches the synchronous model which is very impressive. Although a great time latency, we see at a latency of 550ms that the C+R model outperforms the C+L model when utilizing velocity information. This is one of the only instances in which we have examined C+R outperforming C+L. This is a noteworthy advancement in the research of novel ways to improve radar data, notably for the asynchronous case.

\vspace{0.25em}
Future work includes investigating the effect of training the model from scratch, and not using pretained model. Also, the development of a novel machine learning model to further improve inferring future radar point cloud data will be inquired, perhaps specially tailored for the non-manipulated synchronous datasets.

\section*{Acknowledgements}
This publication has emanated from research conducted with the financial support of Science Foundation Ireland under Grant number 18/CRT/6049. For the purpose of Open Access, the author has applied a CC BY public copyright licence to any Author Accepted Manuscript version arising from this submission.

\appendix

\bibliographystyle{apalike}

\bibliography{imvip}

\end{document}